%% file: pivot_analysis_arxiv.tex
\documentclass[11pt,a4paper]{article}
\usepackage[hyperref]{acl2019}
\usepackage{times}
\usepackage{latexsym}
\usepackage{makecell}
\usepackage{graphicx}
\usepackage{amsmath}
\usepackage{algorithm}
\usepackage{hyperref}
\usepackage{array}
\usepackage{booktabs}
\usepackage{multirow}
\usepackage[noend]{algpseudocode}
\usepackage[draft]{todonotes}

\usepackage{url}

\aclfinalcopy 

\title{Rethinking Text Attribute Transfer: A Lexical Analysis}
 
\author{Yao Fu\Thanks{\;Work done when Yao was an intern at Bytedance AI Lab.} \textsuperscript{1},
Hao Zhou\textsuperscript{2},
Jiaze Chen\textsuperscript{2},
Lei Li\textsuperscript{2}\\
\textsuperscript{1}{Columbia University}\\
\textsuperscript{2}{Bytedance AI Lab}\\
yao.fu@columbia.edu,
zhouhao.nlp@bytedance.com\\
teoyde@gmail.com,
lileilab@bytedance.com}

\date{}

\begin{document}
\maketitle
\begin{abstract}
Text attribute transfer is modifying certain linguistic attributes (e.g. sentiment, style, authorship, etc.) of a sentence and transforming them from one type to another.
In this paper, we aim to analyze and interpret what is changed during the transfer process. 
We start from the observation that in many existing models and datasets, certain words within a sentence play important roles in determining the sentence attribute class.
These words are referred to as \textit{the Pivot Words}. 
Based on these pivot words, we propose a lexical analysis framework, \textit{the Pivot Analysis}, to quantitatively analyze the effects of these words in text attribute classification and transfer. 
We apply this framework to existing datasets and models, and show that: 
(1) the pivot words are strong features for the classification of sentence attributes; 
(2) to change the attribute of a sentence, many datasets only requires to change certain pivot words; 
(3) consequently, many transfer models only perform the lexical-level modification, while leaving higher-level sentence structures unchanged. 
Our work provides an in-depth understanding of linguistic attribute transfer
and further identifies the future requirements and challenges of this task\footnote{Our code can be found at https://github.com/FranxYao/pivot\_analysis}.
\end{abstract}

\section{Introduction}
\label{sec:intro}

\input{010intro.tex}

\section{Background}
\label{sec:background}
\input{020background.tex}

\section{Pivot Words Discovery}
\label{sec:pivot_analysis}
\input{030pivot_analysis.tex}

\section{Analysing Datasets with Pivot Analysis}
\label{sec:datasets}
\input{040dataset.tex}

\section{Analysing Transfer Models with Pivot Analysis}
\label{sec:model}
\input{050model_analysis.tex}

\section{Discussion}
\label{sec:discussion}

\input{060discussion.tex}

\section{Conclusion}
\label{sec:conclusion}

\input{070conclusion.tex}

\section*{Acknowledgments}

We thank the reviewers for their informative reviews. We thank Yansong Feng, Bingfeng Luo, and Zhenxin Fu for the helpful discussions. This work is supported by the China Scholarship Council.

\end{document}

%% file: 010intro.tex
The task of text attribute transfer (or text style transfer
\footnote{Many existing works also call this task style transfer\citep{Fu2018StyleTI}, our work view style as one of the linguistic attributes, and use the term style or attribute according to the context.})
is to transform certain linguistic attributes (sentiment, style, authorship, rhetorical devices, etc.) from one type to another \citep{Ficler2017ControllingLS, Fu2018StyleTI, Hu2017ControllableTG, Li2018DeleteRG, Shen2017StyleTF}. 
The state-of-the-art (SOTA) models have achieved inspiring transfer success rates \citep{Zhao2018AdversariallyRA, Zhang2018SHAPEDSE, Prabhumoye2018StyleTT, Yang2018UnsupervisedTS}.
However, it is still unclear in current literature about what is transferred and what remains to be unchanged during the transfer process. 
To answer this question, we perform an in-depth investigation of the linguistic attribute transfer datasets and models. 

\begin{figure}[t]
\centering
\includegraphics[width=7.5cm]{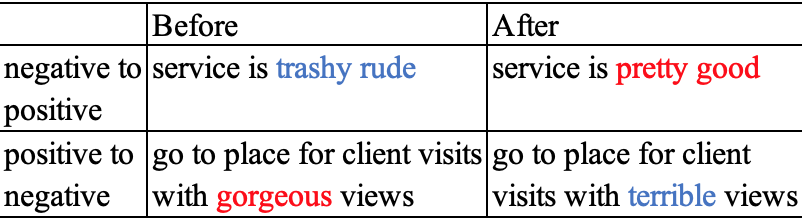}
\caption{\label{fig:intro_examples} Examples of pivot words in sentiment transfer. Certain words are strongly correlated with the sentiment such that a transfer model only need to modify these words to accomplish the transfer task while leaving the higher level sentence structure unchanged.}
\end{figure}

Our investigation starts from a simple observation: in many transfer datasets and models, certain class-related words play very important roles in attribute transfer \citep{Li2018DeleteRG, Prabhumoye2018StyleTT}. Figure \ref{fig:intro_examples} gives a sentiment transfer example from the controllable generation (CG) model \citep{Hu2017ControllableTG} on the Yelp dataset.
In this example, \textit{rude} is strongly related to the {\it negative} sentiment and {\it good} is strongly related to the {\it positive} sentiment, thus simply substituting {\it rude} with {\it good} will transfer the sentence from negative to positive. In this work, We name these words the {\it pivot words} for a class. We use the term \textit{the pivot effect} to refer the effect that certain strong words may be able to determine the class of a sentence. 

Based on the observation of the pivot effect, our research questions are: (1) which words are pivot words and how do they influence the attribute class of a sentence in different datasets? (2) does the model only need to modify the pivot words to perform the attribute transfer or it may change higher-level sentence composationality like syntax? 

\newcolumntype{P}[1]{>{\centering\arraybackslash}p{#1}}

\begin{table*}[t!]
\begin{center}
\small
\begin{tabular}{l | P{1.34cm} P{1.1cm} P{1.1cm} P{1.2cm} P{1.53cm} P{1.53cm} P{1.53cm} P{1.53cm}}
\hline
& Yelp & Amazon & Caption & Paper & Gender & Politics & Reddit & Twitter \\\hline
Source & \citet{Hu2017ControllableTG} & \citet{Li2018DeleteRG} & \citet{Li2018DeleteRG} & \citet{Fu2018StyleTI} & \citet{Prabhumoye2018StyleTT} & \citet{Prabhumoye2018StyleTT} & \citet{Santos2018FightingOL} & \citet{Santos2018FightingOL}  \\\hline
\multirow{2}{*}{Class} & Positive & Positive & Romantic & Academic & Male  & Democratic & Polite & Polite \\
&  Negative & Negative & Humorous & Journalism & Female & Republican & Impolite & Impolite \\\hline
\makecell{Size(train/\\dev/test)} & \makecell{444K/\\63K/126K} & \makecell{554K/\\2K/1K} & \makecell{12K/\\-/1K} & \makecell{392K/\\20K/20K} & \makecell{2M/\\4K/534K}  & \makecell{537K/\\4K/56K} & \makecell{10M/\\19K/47K} & \makecell{3M/\\18K/18K} \\
\hline
\end{tabular}
\end{center}
\caption{\label{tab:dataset} The text attribute transfer datasets we investigate. }
\end{table*}

To answer question (1), we propose the \textit{pivot analysis}, a series of simple yet effective text mining algorithms, to quantitatively examine the pivot effects in different datasets. The basics of the datasets we investigate are listed in Table \ref{tab:dataset}. We first give the algorithm to extract pivot words (Sec \ref{sec:pivot_analysis}). 
We statistically show the stronger the pivot effect is on a dataset, the easier for a model to transfer its sentences. 
To further analyze the fine-grained distributional structure of these pivot words, we propose the \textit{precision-recall histogram} to show to what extent the datasets may be influenced by their pivot words (Sec \ref{ssec:prec_recl_hist}). 

To answer question (2) and discover what is changed during the transfer process, we use the pivot words to analyze the transfer results of two SOTA models: the Controllable Generation (CG) model\citep{Hu2017ControllableTG} and the Cross Alignment (CA) model \citep{Shen2017StyleTF}. We show that although equipped with sophisticated modeling techniques, in many datasets, these models tend to change only a few words and most of these modified words are pivot words. When we mask out the modified words (to eliminate the lexical changes) and compare the Levenshtein string edit distance \citep{levenshtein1966binary} of the sentence stems before and after the transfer, we find out many of the sentence stems are  the same (the distance of the masked sentences equals to 0). This means that in transfer, the model only modifies few pivot words while leaving the syntactical structure of the sentence unchanged (Sec \ref{sec:model}). 

To sum up, we show that: 
(1) in many datasets, words are important features 
in classification and transfer. But still, certain hard cases require a higher level of understanding of the sentence structures. 
(2) SOTA models tend to perform the transfer at the lexical level, the syntax of a sentence is generally unchanged. 
The understanding and modification of higher-level sentence compositionality (syntax trees and dependency graphs) is still a challenging problem.

%% file: 020background.tex

Inspired by the image style transfer task \citep{Gatys2016ImageST, Zhu2017UnpairedIT}, 
the goal of text attribute(style) transfer is to transfer the stylistic attributes of the sentence from one class to another while maintaining the content of the sentence unchanged \citep{Fu2018StyleTI, Ficler2017ControllingLS, Hu2017ControllableTG}.  
Because of the lack of parallel datasets, most models focus on the unpaired transfer.
Although plenty of sophisticated techniques are used in this task, such as adversarial learning \citep{Zhao2018AdversariallyRA, Chen2018AdversarialTG}, latent representations \citep{Li2018DisentangledSA, Dai2019StyleTU, Liu2019RevisionIC}, and reinforcement learning  \citep{Luo2019ADR, Gong2019ReinforcementLB, Xu2018UnpairedST}, there is little discussion about what is changed and what remains unchanged. 

Because of the lack of transparency and interpretability, there is some retrospection on this topic. Such as the definition of text style \citep{Tikhonov2018WhatIW}, and the evaluation metrics \citep{Li2018DeleteRG, Mir2019EvaluatingST}. 
Our proposed pivot analysis aligns with these works and provides a new tool to probe the transfer datasets and models. 
The de facto metrics is to use a pretrained classifier to classify if the transferred sentence is in the target class.
So our pivot analysis starts from the classification task and mines the words with strong predictive performance. 

While many previous works focus on one-to-one transfer, many recent works extend this task to one-to-many transfer \citep{logeswaran2018content, Liao2018QuaSESE, Subramanian2018MultipleAttributeTS}. For simplicity, we focus on the one-to-one setting. But it is also easy to extend the pivot analysis into one-to-many transfer settings.

%% file: 030pivot_analysis.tex
\begin{figure*}[t]
\centering
\includegraphics[width=\textwidth]{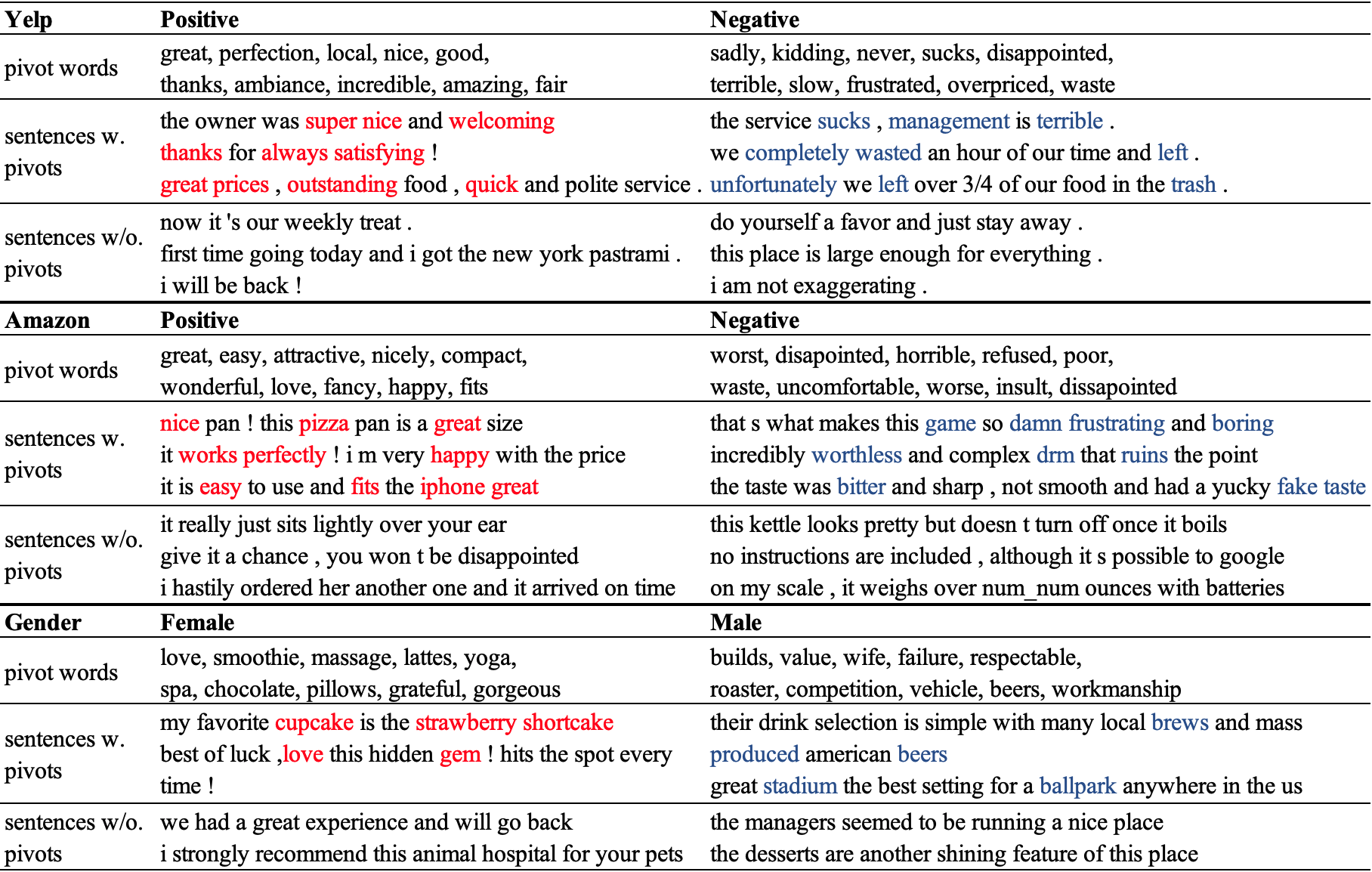}
\caption{\label{fig:pivot_cases} The pivot words and sentence examples in three example datasets. The vocabulary of pivot words is large so we only list typical words. Sentences without pivot words are intuitively harder to classify and transfer.}
\end{figure*}

To study the factors influencing attribute transfer, we start from mining words strongly correlated with the attribute class i.e. pivot words.
Algorithm \ref{alg:pivot_analysis} shows the procedure of mining pivot words. 
This algorithm is based on a simple intuition: if one single word is strong enough to determine the sentence attributes, then when we use the existence of this word to classify the attribute, we should achieve very high precision. 
Consider two extreme examples: when a word only exists in one class, it should achieve 100\% classification precision. When a word exists evenly in two different classes, its precision is 50\%.
The reason we use precision instead of recall or accuracy is that only precision reveals the influence of a single word: suppose the word ``awesome" only exists in 100 positive sentences, and the whole dataset size is 100K. In this case, ``awesome" will have low recall and accuracy, but high precision. 
This algorithm calculates the precision for each word-class pair, and choose pivot words with a predefined threshold $p_0$.

\begin{algorithm}
\small
\caption{\small{Pivot Words Discovery}}\label{alg:pivot_analysis}
\textbf{Input:} The vocabulary $\mathcal{V}$, the sentences $\mathcal{S}$ and the labels $\mathcal{Y}$, the frequency threshold $f_0$, the precision threshold $p_0$ \\ 
\textbf{Output:} The pivot words $\Omega_y$ for each class $y \in \{0, 1\}$. The word-class precision matrix $p(x, y)$
\begin{algorithmic}[1]
\Procedure{Pivot Words Discovery}{}  
\State Balance the dataset by down-sampling the majority class. 
\For{each sentence $s$, each class $y$, and each word $x$ in the vocabulary $\mathcal{V}$ with frequency higher than $f_0$} 
    \State Consider the class of $s$ is $y$ or $1 - y$
    \State Use \textit{the existence of} $x$ to classify: 
    \If{$x$ is in $s$}
        \State Classify $s$ to be $y$
    \Else
        \State Classify $s$ to be $1 - y$
    \EndIf
\EndFor 
\State Calculate the classification precision $p(x, y)$  of word $x$ for label $y$ over all sentences $\mathcal{S}$. 
\If{$p(x, y) > p_0$}
    \State $x$ is a pivot word for class $y$ i.e. $x \in \Omega_y$
\EndIf
\State \textbf{return} $\Omega_y, p(x, y)$
\EndProcedure
\end{algorithmic}
\end{algorithm}

\begin{algorithm}
\small
\caption{\small{The Pivot Classifier}}\label{alg:pivot_classifier}
\textbf{Input:} sentence $s$, the pivot words $\Omega_y$ for class $y \in \{0, 1\}$ \\ 
\textbf{Output:} The class $y(s) $of sentence $s$
\begin{algorithmic}[1]
\Procedure{Pivot Classification}{} 
\State View $s$ as bag of words 
\State For each $y \in \{0, 1\}$, calculate $s_y = ||s 	\cap \Omega_y||$ 
\State Predict the class of $s$ to be $y(s) = \text{argmax}_y\{s_y\}$. Break tie randomly.
\State \textbf{return} $y(s)$
\EndProcedure
\end{algorithmic}
\end{algorithm}

For simplicity, we only consider binary classification in Algorithm \ref{alg:pivot_analysis}, 
but one could easily extend it to multi-class settings. 
Also, we only consider unigrams(words), while it is also straightforward to extend it to ngrams. 
In practice, we find the unigram version performs quite good, as is shown in Table \ref{tab:classification_retults}. 
As for the parameters in the algorithm,
the precision threshold $p_0$ controls the confidence of a word to be a pivot, and the occurrence threshold $f_0$ prevents overfitting. 
We tune these parameters based on the classification performance on the validation set. 
Specifically, to get better classification performance, $f_0$ and $p_0$ should be lower to allow more vote (e.g. $f_0 \le 10, p_0 \in [0.5, 0.7]$).
To get more confidence and filter out stronger pivot words, $f_0$ and $p_0$ should be higher (e.g. $f_0 \ge 100, p_0 \ge 0.7$).

\begin{table*}[t!]
\begin{center}
\small
\begin{tabular}{l|l l l l l l l l}
\hline 
\bf Validation & Yelp & Amazon & Caption & Gender & Paper & Politics & Reddit & Twitter \\\hline
Pivot & 88.00 & 75.85 & -  & 72.02  & 97.82 & 98.32 & 90.00  & 85.25 \\
Logistic & 91.83 & 76.75 & - & 72.77 & 98.39 & 99.82
 & 98.05 & 98.20
\\
CNN & 92.87 & 77.93 & - & 74.20 & 98.36 & 98.85 & 99.45 & 99.55 \\ \hline \hline
\bf Test & Yelp & Amazon & Caption & Gender & Paper & Politics & Reddit & Twitter \\\hline
Pivot & 88.35 & 73.30 & 69.20 & 71.91 & 98.07 & 94.60 & 91.15 & 85.05
 \\
Logistic & 91.97 & 73.80 & 75.20 & 72.91 & 98.67 & 96.83 & 98.05 & 98.05 \\
CNN & 92.96 & 75.80 & 76.10 & 74.29 & 98.66 & 87.91 & 99.65 & 99.45 \\
\hline
\end{tabular}
\end{center}
\caption{\label{tab:classification_retults} Classification accuracy. The voting based pivot classifier is a strong classification baseline compared with the state of art CNN classifier, indicating that in many datasets, words are strong features for class labels.}
\end{table*}

Figure \ref{fig:pivot_cases} shows the mined pivot words in different datasets.
For sentences that contain pivot words, it is clear that these words are strong features for classification. 
Intuitively, to transfer the class of these sentences, one could directly modify these words.
But there are also cases that contain no pivot words, e.g. \textit{i will be back} in the Yelp dataset. 
To modify the sentiment of these sentences, a model needs to understand a broader context and common sense. 
In general, the existence of pivot words gives us a method to understand in attribute transfer, what cases are easier and what cases are more difficult. 

The intuition that the existence of single words is enough to determine the linguistic attribute does not necessarily hold on all datasets. 
But empirically, we find out many transfer datasets tend to contain strong pivot words (Figure \ref{fig:prec_recl}). 
One could compare our pivot analysis with other methods that mine the word importance, such as the weights of a logistic classifier, or more sophisticated Bayesian methods like the log-odds ratio informative Dirichlet prior \citep{Monroe2008FightinWL}. Our method is more straightforward and interpretable. 
We further develop this method as a simple yet strong classification baseline to indicate the transfer difficulty of different datasets and use the pivot words as a tool to analyze, interpret, and visualize the text attribute transfer models.

%% file: 040dataset.tex
In this section, we use the pivot words to analyze the transfer datasets. 
We first reveal the mechanisms of how pivot words affect classification and transfer by using the pivot words as the classification boundary. 
Then we use the precision-recall histogram to demonstrate the distributional structure of the pivot words in different portions of the datasets.

\subsection{The Pivot Classifier}
\label{ssec:pivot_classifier}

Algorithm \ref{alg:pivot_classifier} gives a simple method to classify a sentence based on the pivot words output from Algorithm \ref{alg:pivot_analysis}.
This is essentially a voting based classifier. 
This classifier holds strong independence assumption that the label of a sentence is only related to the bag of words, but ignore the word orders. 
This is to say, the decision boundary only stays at the lexical level, and does not go to the syntax level. 
Then it counts the pivot words of different classes contained by the sentence and predicts the label to be one of the largest pivot words overlap. 
Intuitively, this algorithm classifies a sentence only based on the existence of strong attribute-related words. 

\begin{figure}[t]
\centering
\includegraphics[width=7.7cm]{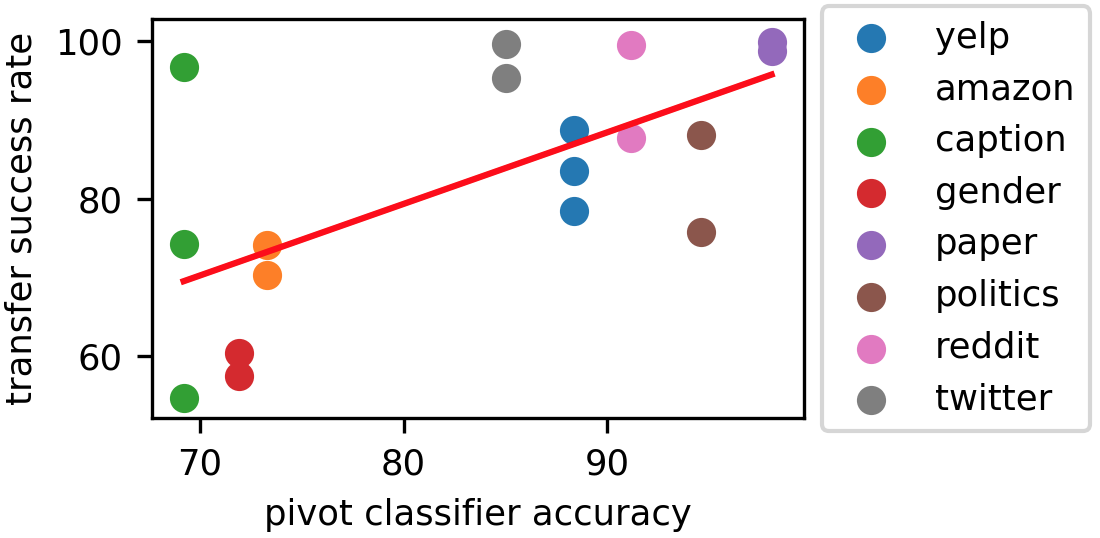}
\caption{\label{fig:pivot_transfer} Pivot classification accuracy v.s. transfer success rate
 (correlation = 0.64, p-value = 0.003). The stronger the pivot effect is, the easier to transfer.}
\end{figure}

\begin{figure}[t]
\centering
\includegraphics[width=7.7cm]{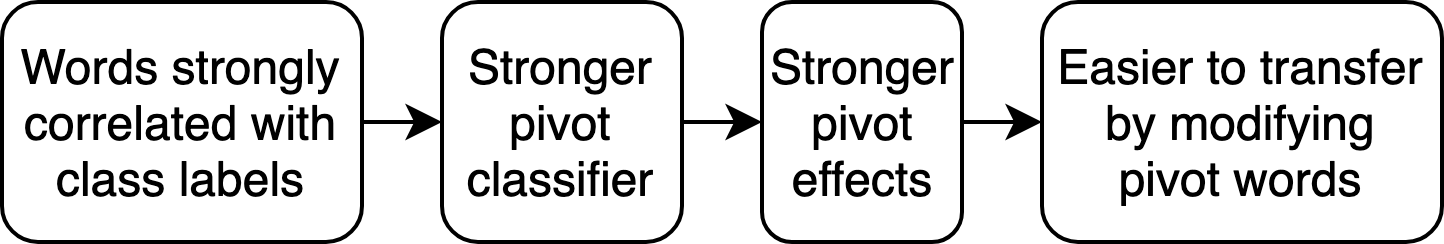}
\caption{\label{fig:pivot_mechanism} The mechanism of the pivot effect on classification and transfer.}
\end{figure}

\begin{figure*}[t]
\centering
\includegraphics[width=\textwidth]{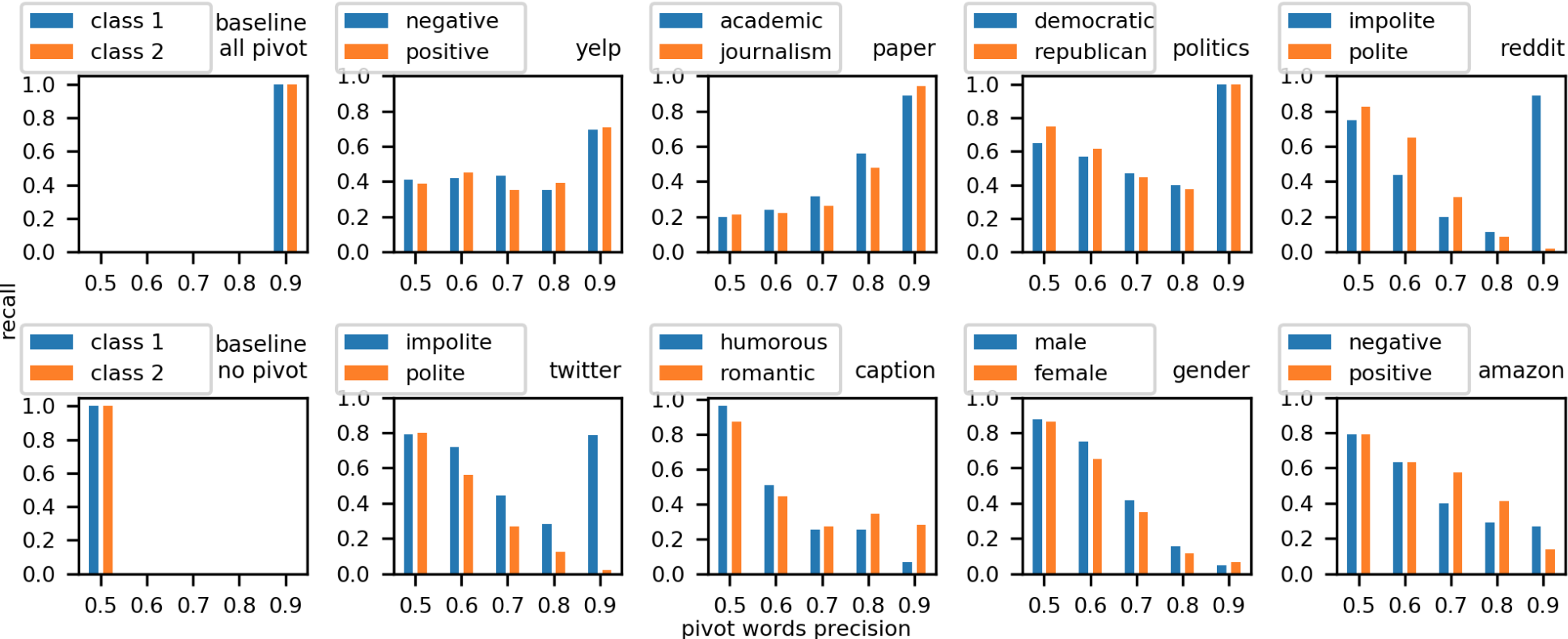}
\caption{\label{fig:prec_recl} The precision-recall histogram. The high right bars in Yelp, Paper, Politics, Reddit, and Twitter datasets reveal the existence of strong pivot words,  Each bar at location $(x, y)$ should be interpreted as: if use pivot words with precision $x$ to classify the sentence, the recall will be $y$. The higher the right bars are, the more sentences can be classified by words accurately, the stronger the pivot effect is, the easier to transfer. The baseline cases where the dataset is full of/ has no pivot words are show on the left.}
\end{figure*}

The pivot classifier is a simple yet strong classification baseline, as is shown in Table \ref{tab:classification_retults}.
We use it to study different datasets and compare it with (1) a logistic classifier, (2) a SOTA CNN classifier \citep{kim2014convolutional}.
We have balanced the test sets so the random baseline is 50\%. 
This voting based classifier achieves comparable performance with the two models in 4 datasets (Amazon, Gender, Paper, Politics), and only loses small margins in 2 datasets (Yelp, Caption). 
Although the independence assumption from our pivot classifier does not necessarily hold for all datasets, empirically it performs very well. 
This means that these pivot words are a meaningful approximation of the true decision boundary. 

If the decision boundary of a linguistic attribute stays at the lexical level, then one could cross the boundary by simply substituting the pivot words of one class to another, thus achieving text class transfer. 
Intuitively, the more pivot words a dataset contains, the stronger the pivot effect is, the easier for the pivot classifier to classify, and the easier to transfer the attribute. 
This intuition is demonstrated in Figure \ref{fig:pivot_transfer}. 
The pivot effect (shown by pivot classification accuracy) and the transfer difficulty (shown by the transfer success rate reported from previous models) has a strong positive correlation and is statistically significant. 
This mechanism is demonstrated in Figure \ref{fig:pivot_mechanism}.
The stronger the pivot effect is, the easier to transfer. 

\subsection{The Precision-Recall Histogram}
\label{ssec:prec_recl_hist}

Now we go one step further to reveal how the pivot effect distributes in different portions of the datasets. 
We propose a new tool, \textit{the precision-recall} histogram based on the results from Algorithm \ref{alg:pivot_analysis} and \ref{alg:pivot_classifier}.
As is shown in Algorithm \ref{alg:prec_recl}, 
essentially, this algorithm use pivot words with different level of confidence (precision) to classify the dataset, and output the recall. 
For better visualization, we set the precision interval gap to be 0.1, but it is also possible to use smaller or larger gaps. 
It is also important to balance the dataset in Algorithm \ref{alg:pivot_analysis} to make the baseline precision 0.5. 

The histogram for all datasets gives a fine-grained illustration of the pivot effect (Figure \ref{fig:prec_recl}). 
We first look at the two baseline cases: a dataset with no pivot words, and a dataset full of pivots. 
If a dataset is full of pivots, i.e. the vocabulary of the two classes have no overlap, then all words should have precision 1.0 and they should achieve 1.0 recall, so the right-most bars are the highest. 
If a dataset has no pivot words, i.e. all words are distributed evenly in two classes, then all words have precision 0.5 and they should achieve 1.0 recall, so the left-most bars are the highest.
The higher the right bars are, the stronger the pivot effect is. 

\begin{algorithm}
\small
\caption{\small{The Precision-Recall Histogram}}\label{alg:prec_recl}
\textbf{Input:} The sentences $\mathcal{S}$, the labels $\mathcal{Y}$, the pivot words for each class $\Omega_{y}, y\in \mathcal{Y}$, the precision matrix $p(x, y), x \in \mathcal{V}, y \in \mathcal{Y}$ \\ 
\textbf{Output:} The precision-recall histogram
\begin{algorithmic}[1]
\Procedure{The Precision-Recall Histogram}{} 
\For{The precision range pair $(p_i, p_{i + 1}) \in [(0.5, 0.6), (0.6, 0.7) ... (0.9, 1.0)]$}
\State For each class $y$, gather all pivot words of the precision in the given range: $\Omega_y^{(i)} = {x: p(x, y) \in [p_{i}, p_{i + 1}]}$
\State Use $\Omega_y^{(i)}$ to form a pivot classifier and  classify the dataset $\mathcal{S}$. Calculate the recall $r_i$.
\State Store ($p_i, r_i$)
\EndFor
\State \textbf{return} The list of $(p_i, r_i)$
\EndProcedure
\end{algorithmic}
\end{algorithm}

The histograms of the datasets are somewhere between the two baseline cases. 
Generally, we see two different shape distributions. 
In the Yelp, Paper, Politics, Reddit, and Twitter datasets, the right-most bars are the highest, meaning that in these datasets, strong pivot words exist in a large portion of the dataset. 
These are close to the all-pivot baseline.
Specially, we see that in the Reddit and Twitter dataset, the pivot effect only exists in the \textit{impolite} class, while in other datasets, the pivot effect exists in both classes. 
Note that this phenomenon cannot be discovered simply from the overall classification accuracy. 
After manual inspection, we find out since the attribute of these two datasets is politeness, the pivot words for the impolite class are the common  swearwords in English. 
These words dominate the impolite sentences. 

In the Caption, Gender, and Amazon dataset, we see a decreasing height from left to right, indicating a weaker pivot effect. 
Highest bars exist in the 0.5 precision bars, meaning that for each class, most of them can be classified by 0.5 precision (= random guessing).
This is close to the no-pivot baseline.
The high-precision words still exist, but they cannot dominate the whole class. 
In conclusion, the precision-recall histograms give a structural examination for each class. 
The existence of pivots and the determination power of pivots differ from class to class, and from datasets to datasets.

%% file: 050model_analysis.tex
\begin{table}[t!]
\begin{center}
\small
\begin{tabular}{l|lll}
\hline 
 & Yelp  & Amazon & Gender\\\hline
CG - \# modified & 1.66 & 0.56 & 0.79\\ 
$\;\;\;\;\;$ - percentage & 18\% & 4\% & 5\% \\ \hline
CA - \# modified & 1.61 & 3.54 & 5.60 \\ 
$\;\;\;\;\;$ - percentage & 18\% & 23\% & 33\% \\ \hline
sentence length & 8.89 & 14.82 & 17.01 \\
\hline
\end{tabular}
\end{center}
\caption{\label{tab:num_modified_words} Average number of modified words and their percentage in the sentence length. The transfer models tend to modify only a few attribute-related words.}
\end{table}

\begin{table}[t!]
\begin{center}
\small
\begin{tabular}{l|lll}
\hline 
 & Yelp  & Amazon & Gender\\\hline
CG & 91.25 & 94.77 & 94.17 \\
CA & 72.33 & 74.04 & 56.09 \\ 
\hline
\end{tabular}
\end{center}
\caption{\label{tab:percent_modified_pivot} Percentage of modified words that are pivot words. A large portion of the modified words are pivots.}
\end{table}

In this section, we aim to analyze what is changed and what remains in linguistic attribute transfer systems. 
We perform our experiments from two perspectives: the lexical structures, and the syntactical structures. 
For the lexical structures, we show what words are modified by the transfer model. 
For the syntactical structures, we mask out the modified pivot words and compare the resulting sentence stems.

We use the two most common SOTA models, the Controllable Generation (CG) model from \citet{Hu2017ControllableTG}, and the Cross Aligned Autoencoder (CA) model from \citet{Shen2017StyleTF}. 
The CG model uses a conditional VAE with style-discriminator and trained with a wake-sleep algorithm. 
The CA model uses a cross-alignment mechanism to guide the transfer process. 
These are two strong models in many datasets compared to many other models.
We direct the readers to the original papers for more details.

We test the models on three datasets: Yelp, Amazon, and Gender.
The Yelp dataset is the most widely used benchmark in the text style transfer task. 
As is shown in the previous sections, it exists strong pivot effects. 
There are many sentiment words in this dataset. 
For the Amazon and the Gender dataset, there is less pivot effect. 
So our experiments give a minimum cover of different types of datasets. 
We use the released implementation for our experiments
\footnote{The CG model: \url{https://github.com/asyml/texar/tree/master/examples/text_style_transfer}

the CA model: \url{https://github.com/shentianxiao/language-style-transfer}}. 
All hyper-parameters are followed by their official instructions.
Both models are trained until the simultaneous convergence of the reconstruction loss and the adversarial loss. 
We refer the readers to the implementation repositories for more details.

\begin{table*}[t!]
\begin{center}
\small
\begin{tabular}{l|llllllll}
\hline 
CG & 0  & 1 & 2 & 3 & 4 & 5 & 6 & $>$6 \\\hline
Yelp & 74.65 & 5.05 & 10.68 & 5.71 & 1.84 & 0.72 & 0.66 & 0.69 \\ 
Amazon & 94.20 & 0.00 & 0.90 & 4.00 & 0.80 & 0.10 & 0.00 & 0.00 \\ 
Gender & 90.96 & 0.04 & 6.60 & 0.45 & 0.89 & 0.43 & 0.16 & 0.48 \\ \hline 
CA & 0  & 1 & 2 & 3 & 4 & 5 & 6 & $>$6 \\\hline
Yelp & 41.30 & 1.98 & 13.63 & 10.01 & 8.76 & 7.49 & 6.17 & 10.66\\ 
Amazon & 37.60 & 1.85 & 9.95 & 9.25 & 6.15 & 6.15 & 4.65 & 24.40\\ 
Gender & 37.89 & 0.27 & 2.27 & 3.36 & 1.60 & 1.40 & 2.03 & 51.18\\ \hline
\end{tabular}
\end{center}
\caption{\label{tab:edit_dist_distribution} Masked edit distance percentage distributions. For the CG model, in most of the cases($>74$\%), the masked edit distance is 0, meaning that only few words are changed while the sentence structures are exactly the same. For the CA model, still a large portion of the sentence structures are unchanged ($>37$\%)}
\end{table*}

\begin{figure*}[h]
\centering
\includegraphics[width=0.65\textwidth]{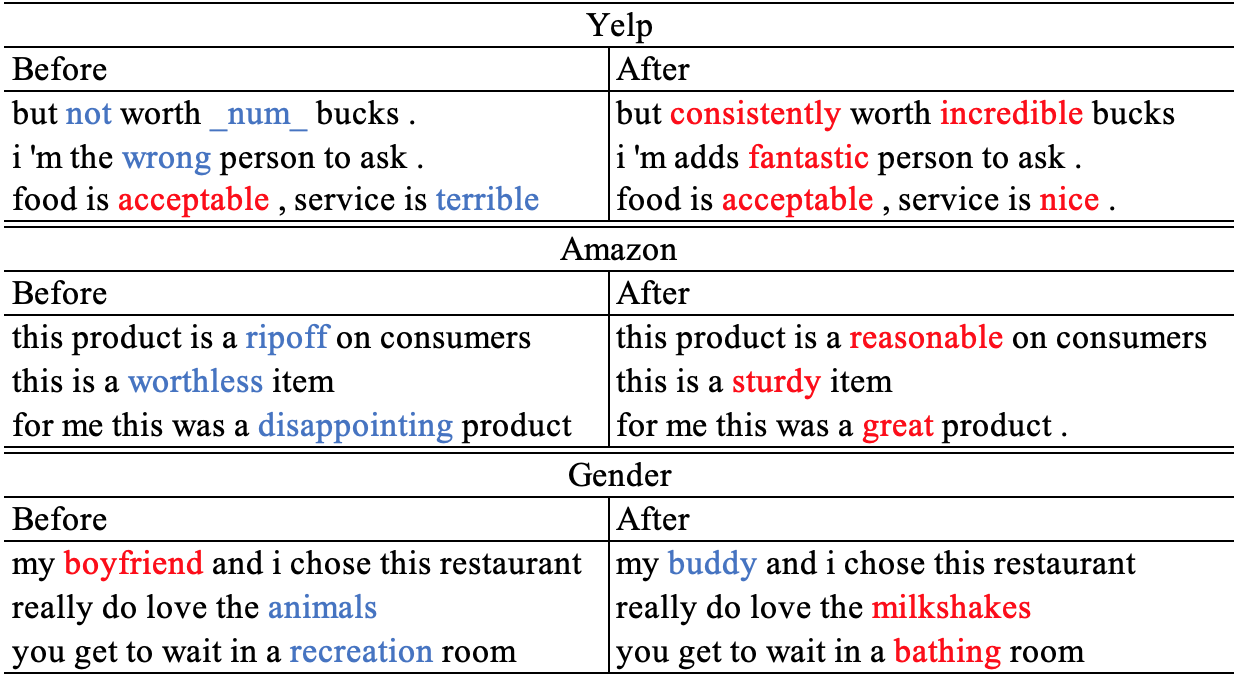}
\caption{\label{fig:transfer_cases} The transfer cases. Many of the transfered words are pivot words. The model tend to  transfer only a few words while leaving the higher level sentence structure unchanged.}
\end{figure*}

\begin{figure}[h]
\centering
\includegraphics[width=7cm]{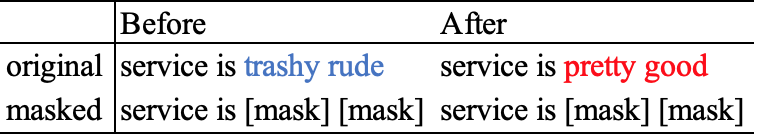}
\caption{\label{fig:masked_example} An example of the masked sentences. Edit distance = 0 after masking.}
\end{figure}

\begin{table}[h]
\begin{center}
\small
\begin{tabular}{l|lll}
\hline 
 & Yelp  & Amazon & Gender\\\hline
CG - distance & 0.65 & 0.17 & 0.26 \\
$\;\;\;\;\;$ - percentage & 7.3\% & 1.4\% & 1.5\% \\ \hline
CA - distance & 2.63 & 4.56 & 9.95 \\ 
$\;\;\;\;\;$ - percentage & 29\% & 31\% & 58\% \\ \hline
sentence length & 8.89 & 14.82 & 17.01 \\
\hline
\end{tabular}
\end{center}
\caption{\label{tab:edit_dist} Edit distance after masking out the pivot words. In the CG model, only words are modified, while the higher-level sentence structures remain to be the same. For the CA model, it tries to modify more sentence structures.}
\end{table}

\subsection{Lexical Structures}

We show that the two models tend to modify only a few words in a given sentence, and a large portion of these words are pivot words.
The results are shown in Table \ref{tab:num_modified_words} and \ref{tab:percent_modified_pivot}.
On the Yelp dataset, the CG model and the CA model only modify 1.66 and 1.61 words on average. 
The portion of pivot words is 91\% and 72\% respectively. 
This means on this dataset, both two models focus on word substitutions to change the sentence style. 
On the Amazon and the Gender dataset, the models take different transfer strategies. 
For the CG model, it concentrates on fewer words to modify (0.56 on Amazon and 0.79 on Gender). 
For the CA model, it tends to modify more words (3.54 on Amazon and 5.60 on Gender). 
Still, both models tend to modify the pivot words for class transfer.
In general, a small portion of the sentences are modified ($< 30\%$ approximately), and a large portion of the modified words are pivots ($> 60\%$ approximately). 


\subsection{Syntactic Structures}

If we eliminate the lexical differences by masking out the modified words, what is changed in the resulting sentence stems? 
We use the Levenshtein string edit distance \citep{levenshtein1966binary} to measure the distances of the masked sentences as an approximation to the distances of syntactic structures. 
Figure \ref{fig:masked_example} gives an example of masked sentences. 
One could also consider more sophisticated metrics to measure the syntactic distances with parsing trees \citep{shen2018straight, zhang1989simple}. 
Here we use the string edit distance for simplicity. 
In practice, it is informative enough to demonstrate the change of sentence structures. 

Table \ref{tab:edit_dist} shows the edit distances after masking the pivot words. 
We see clear differences between the two models.
For the CG model, it barely changes the sentence structures (0.1+ distances). 
This indicates that it takes the strategy to focus more on the substitution of pivot words. 
For the CA model, it takes the strategy that not only to modify the words, but also a portion of the sentence structures. 
We see a moderate percentage of the sentence structure modified on the Yelp and Amazon dataset (about 30\%), and a large syntactic modification (58\%) on the Gender dataset. 
Compared with the CG model, the CA model tries to modify the sentences more radically. 

To show a fine-grained distribution of the distances among different cases, we list the distribution statistics in Table \ref{tab:edit_dist_distribution}. 
We see that for the CG model, most of the cases $>74$\%) sentence stems are unchanged. 
For the CA model, although its average edit distance is larger, in a large portion of the cases ($>37$\%), the distance is still 0. In conclusion, both models tend to retain the sentence structures in a large portion of the datasets.

\subsection{Qualitative Analysis}

Now we examine the transfer cases qualitatively in Figure \ref{fig:transfer_cases}. 
These are cases from the CG model on the three datasets. 
The pivot words are highlighted. 
When the model tries to change the class of a sentence, it first identifies the pivot words, then substitutes them with the pivots from another class. 
If we mask out the highlighted pivot words, the resulting sentence stems are the same, indicating that the syntactic structures remain unchanged. 
Although this is not all the case, the models tend to focus on words in a large portion of the datasets.

%% file: 060discussion.tex
\paragraph{Implications:} Our pivot classifier reveals that to a certain extent, in many transfer datasets, the decision boundary stays at the lexical level. 
Consequently, to cross the boundary and transfer the text class, many instances in the dataset only requires to modify certain pivot words. 
But still, there are cases with no pivot words.
The decision boundary in such cases is higher than the word level. 
To transfer these cases, the model needs a deeper understanding of the sentence structures, which may include syntax, semantics, and common sense (Figure \ref{fig:pivot_cases}).

\paragraph{Considerations:} In our experiments, we find out the two models are both quite unstable during training. 
The balance between the reconstruction loss and the adversarial loss will significantly influence the convergence point. 
Our pivot analysis framework requires the model to converge to a meaningful local optimum with reasonable content preservation and transfer strength at the same time\citep{Fu2018StyleTI}. 
For our pivot algorithms, it is important to balance the datasets (both training and testing) for a reasonable precision baseline(0.5). 
Our algorithm is mostly sensitive to the precision threshold $p_0$ i.e. the confidence of how \textit{pivot} a word is. 
We tune this parameter based on the development set performance. 

\paragraph{Limitations:} All of our pivot algorithms stays at \textit{the lexical level}. 
These algorithms hold strong Independence assumption that the class of a sentence is independent of the order of words. 
So this method may not be able to capture certain linguistical phenomenons, such as anastrophe
\footnote{To change the order of certain words}. 
One could also consider an extreme example where the pivot analysis \textit{does not work}: suppose we have a corpus of sentences, we label all of them to be 0, then we \textit{reverse all sentences}, and label the reversed sentences to be 1. 
In this dataset, both classes share the same vocabulary, and the precision of any word will be 0.5. 
This is an example where only the order determines the class.
Further, in our work, we only consider  lexical changes, and do not consider other issues with regard to more rigorous definition of linguistic style\citep{Tikhonov2018WhatIW}, the evaluation metrics \citep{Mir2019EvaluatingST}, and the causality in text classification\citep{wood2018challenges}. 
These topics will be the future directions.

%% file: 070conclusion.tex
In this work, we present \textit{the Pivot Analysis}, a lexical analysis framework for the examination and inspection of text style transfer datasets and models. 
This analysis framework consists of three text mining algorithms, \textit{pivot words discovery, the pivot classifier, and the precision-recall histograms}. 
With these algorithms, we reveal what are the important words that influence the class of a sentence, how these words are distributed in a dataset, the mechanisms through which these words interact with a transfer model, and how the models perform the transfer. 
Our method serves as a probe for the transparency and the interpretability of the datasets and the transfer models.
We show that a large portion of the transfer cases stays at the lexical level, while the syntactic structures are unchanged.  

Since our methods stay at the lexical level, it has its own limitations in understanding higher-level sentence compositionality. 
These limitations are also shared by the SOTA transfer models: to understand the syntax and semantics (i.e. the structures of the sentence), and the common sense (i.e. the background and implications of the surface words). 
These limitations are also directions for future challenges. 
In the future, we need to use better inductive bias and use more powerful models towards higher-level sentence compositionality. 